\documentclass[letterpaper]{article} 
\usepackage{aaai25}  
\usepackage{times}  
\usepackage{helvet}  
\usepackage{courier}  
\usepackage[hyphens]{url}  
\usepackage{graphicx} 
\usepackage{natbib}  
\usepackage{caption} 
\usepackage{algorithm}
\usepackage{algorithmic}
\usepackage{newfloat}
\usepackage{listings}
\usepackage{amssymb}
\usepackage{amsmath}  
\usepackage{booktabs} 
\usepackage{multirow} 
\usepackage{float}
\usepackage{newfloat}
\usepackage{listings}
\DeclareCaptionStyle{ruled}{labelfont=normalfont,labelsep=colon,strut=off} 

\floatstyle{ruled}
\newfloat{listing}{tb}{lst}{}
\floatname{listing}{Listing}

\title{Mesoscopic Insights: Orchestrating Multi-scale \& Hybrid Architecture for Image Manipulation Localization}
\author{Xuekang Zhu{\textsuperscript{\rm 1,2}\equalcontrib},
Xiaochen Ma{\textsuperscript{\rm 3}\equalcontrib}, Lei Su{\textsuperscript{\rm 1,2}}, Zhuohang Jiang{\textsuperscript{\rm 4}}, Bo Du{\textsuperscript{\rm 1,2}}, Xiwen Wang{\textsuperscript{\rm 1,2}}, Zeyu Lei{\textsuperscript{\rm 1,2}}, Wentao Feng{\textsuperscript{\rm 1,2}}, Chi-Man Pun{\textsuperscript{\rm 5}}, Jizhe Zhou{\textsuperscript{\rm 1,2}\thanks{Corresponding author: Jizhe Zhou {jzzhou@scu.edu.cn}}
}}
\affiliations{
\textsuperscript{\rm 1}College of Computer Science, Sichuan University \\
\textsuperscript{\rm 2}Engineering Research Center of Machine Learning and Industry  Intelligence, Ministry of Education of China \\
\textsuperscript{\rm 3}Mohamed Bin Zayed University of Artificial Intelligence\\
\textsuperscript{\rm 4}The Hong Kong Polytechnic University\\
\textsuperscript{\rm 5}Computer and Information Science, Faculty of Science and Technology, University of Macau
}

\begin{document}

\maketitle

\begin{abstract}
The mesoscopic level serves as a bridge between the macroscopic and microscopic worlds, addressing gaps overlooked by both. Image manipulation localization (IML), a crucial technique to pursue truth from fake images, has long relied on low-level (microscopic-level) traces. However, in practice, most tampering aims to deceive the audience by altering image semantics. As a result, manipulation commonly occurs at the object level (macroscopic level), which is equally important as microscopic traces. Therefore, integrating these two levels into the mesoscopic level presents a new perspective for IML research. Inspired by this, our paper explores how to simultaneously construct mesoscopic representations of micro and macro information for IML  and introduces the Mesorch architecture to orchestrate both. Specifically, this architecture \textbf{i)} combines Transformers and CNNs in parallel, with Transformers extracting macro information and CNNs capturing micro details, and \textbf{ii)} explores across different scales, assessing micro and macro information seamlessly. Additionally, based on the Mesorch architecture, the paper introduces two baseline models aimed at solving IML tasks through mesoscopic representation. Extensive experiments across four datasets have demonstrated that our models surpass the current state-of-the-art in terms of performance, computational complexity, and robustness. 
\begin{links}
    \link{Code}{https://github.com/scu-zjz/Mesorch}
\end{links}

\end{abstract}

\section{Introduction}
The mesoscopic system exists between macroscopic and microscopic scales. Objects at the mesoscopic scale are large enough to exhibit macroscopic properties, yet they also display interference phenomena related to quantum mechanical phases, similar to microscopic systems. This duality is why it is termed ``mesoscopic"---\textit{Yoseph lmry}.

The rapid advancement of multimedia tampering techniques has made detecting and localizing image manipulation more challenging. The ease of creating realistic fakes has fueled tampering incidents and misinformation, highlighting the need for effective forensic methods.
\begin{figure}[t]
    \centering
    \includegraphics[width=0.98\columnwidth]{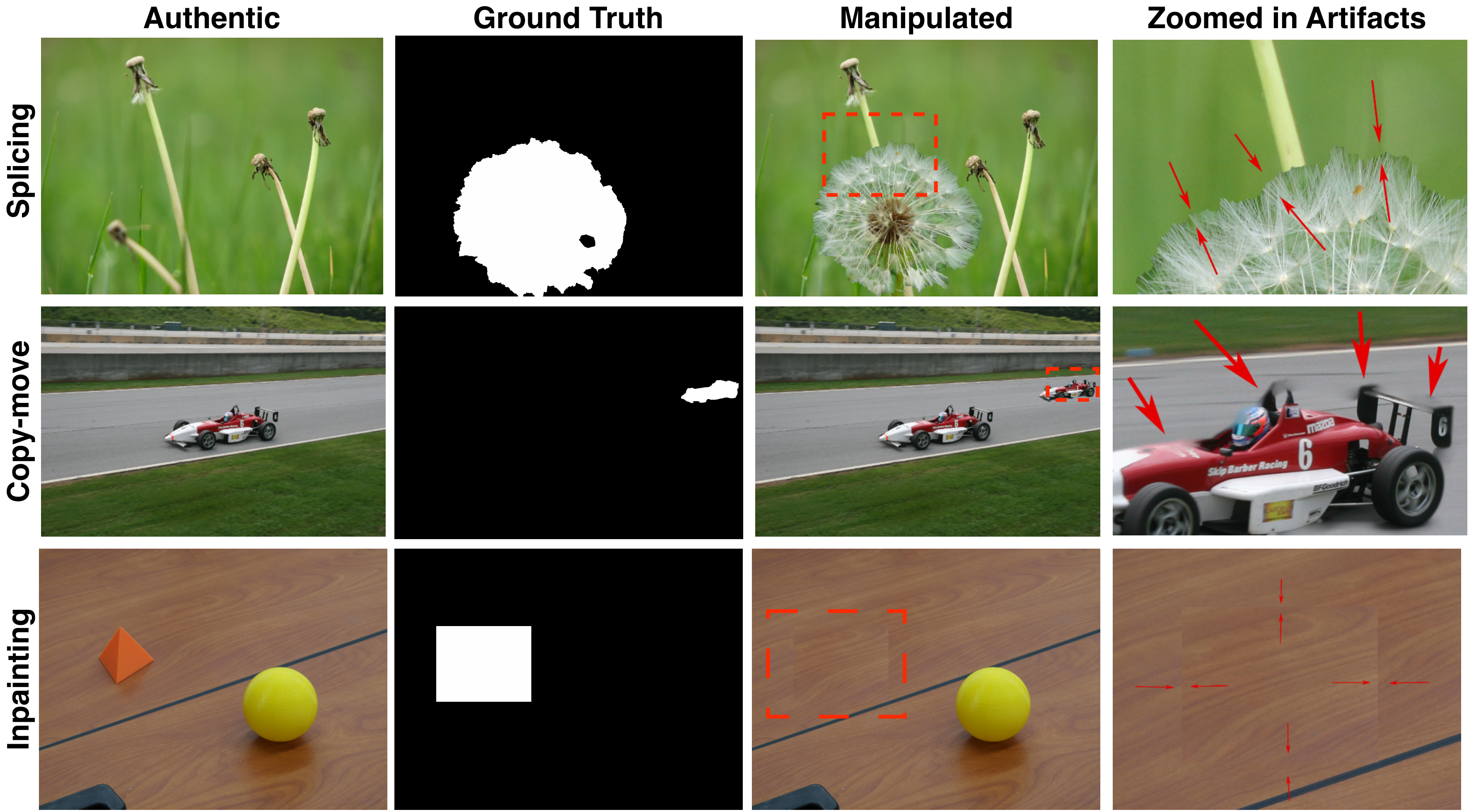}
    \caption{Example of artifacts in three types of tampering.  The red dashed box in the third column represents the range of the zoomed-in area in the fourth column.  Red arrows in the fourth column point to artifacts that are considered tampering traces.}
    \label{fig:intro_example}
\end{figure}

In this context, as illustrated in Figure \ref{fig:intro_example}, existing tampering techniques can be broadly categorized into three types~\cite{pun2015image,bi2016multi,Mantra_2019, Verdoliva_2020,wei2023secondary}: Splicing (combining parts of different images to create a new one), Copy-move (copying and pasting regions within the same image), and Inpainting (removing and then filling an area with plausible content). Although intricate manual tampering can be imperceptible to the human eye, each type of manipulation still leaves detectable traces at the low level (microscopic level). Therefore, most current techniques in image manipulation localization (IML) consider it a microscopic level task aimed at capturing these tampering traces (artifacts) by extracting microscopic features such as image RGB noise~\cite{SRM_2018, Bayar_2018}, edge signals~\cite{GSR_Net_2020, MVSS_2021}, or high-frequency features~\cite{HPFCN_2019, objectformer_2022}. These microscopic features are generally effective in revealing artifacts and localizing tampered areas.

\begin{figure}[t]
    \centering
    \includegraphics[width=\linewidth]{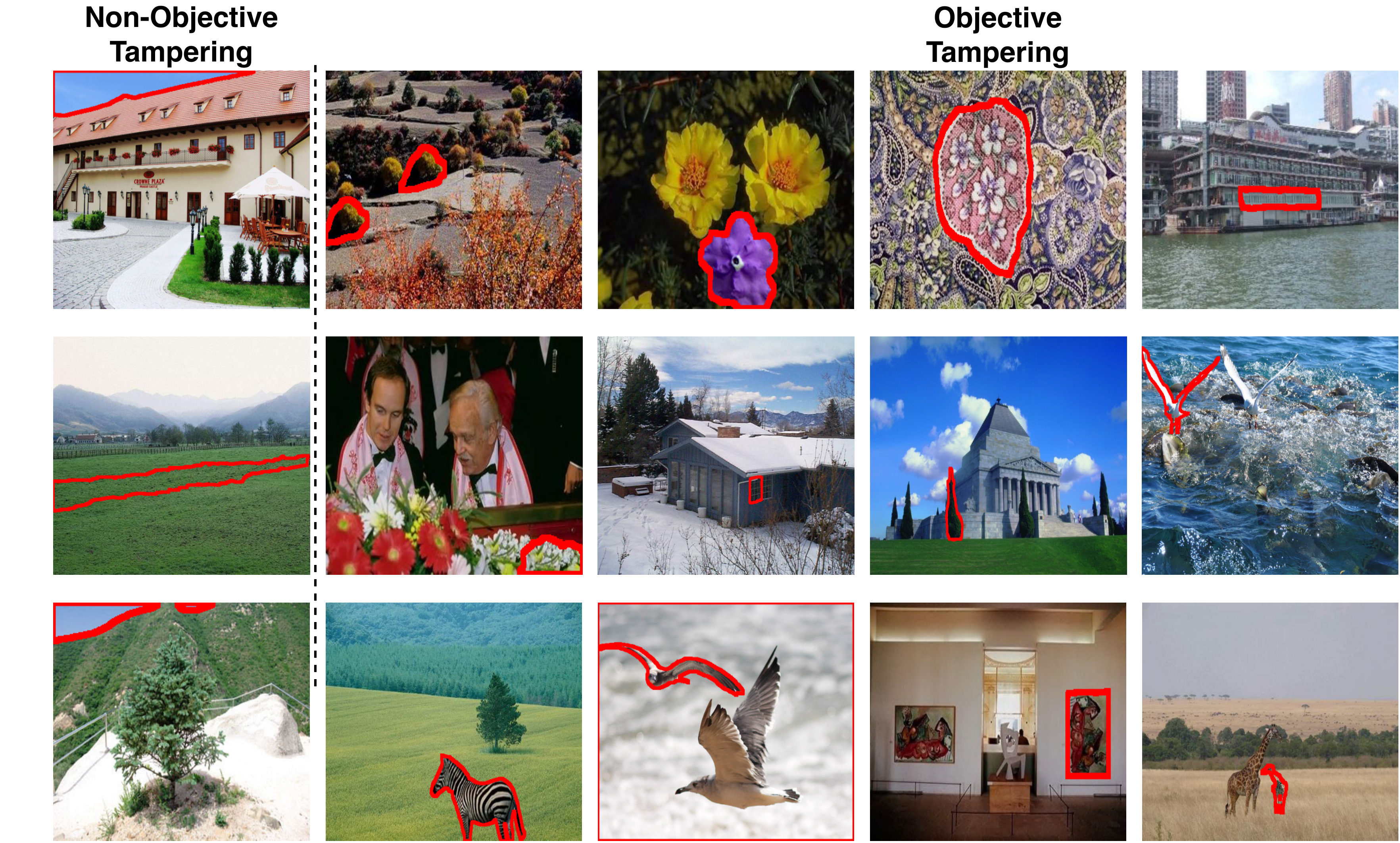}
    \caption{ Random samples from the CASIAv2 dataset. The red line marks the clear boundary of the tampered area. The first column shows tampering that is entirely unrelated to objects, while the other four columns show object-related tampering.}
    \label{fig:intro_object}
\end{figure}

However, as depicted in Figure \ref{fig:intro_object}, the majority of tampering is typically aimed at deceiving the audience by altering or obscuring the semantics of images. For instance, we observe that in the CASIAv2 dataset, around 80\% of the samples are related to the manipulation of objects.  Further, the possibility of an object being tampered with will change according to its contribution to the overall semantics of the image. For example, humans and animals in the foreground are more likely to be tampered with than trees and mountains in the background. Therefore, we argue that understanding the object-level (macroscopic-level) semantics is crucial for identifying manipulated regions and delineating suspicious areas. Nonetheless, macroscopic-level semantics alone is insufficient for generating tampering masks, as it lacks necessary details to detect intricate artifacts.

Therefore, figuring out a path to integrate both micro and macro information into the mesoscopic level could be a new solution to improve IML research. 
DiffForensics~\cite{diffforensics2024} defined that artifacts exist at the mesoscopic level, which means that they exist between the microscopic and macroscopic levels and share the characteristics of both. However, their research did not delve deeper into how to technically characterize this level. To address this, based on our previous in-depth analysis, we define capturing artifacts at the mesoscopic level by simultaneously capturing microscopic features and extracting macroscopic semantics as a new paradigm for IML tasks. Derived from this paradigm, we introduce the Mesorch (\textbf{Meso}scopic-\textbf{Orch}estration) architecture, which employs parallel en

         r and decoder structures specifically designed to represent the mesoscopic level, enabling more effective orchestration between both levels and thus more precisely capturing artifacts at the mesoscopic level.

In the encoder stage, current research~\cite{part2016,interpreting2018,hrformer2021} has shown that CNN and Transformer models excel in processing microscopic features and macroscopic semantics, respectively. However, most existing IML methods~\cite{Mantra_2019,SPAN_2020,trufor2023} still rely exclusively on either CNNs or Transformers for decision-making. Although some models, such as ObjectFormer~\cite{objectformer_2022}, have recognized the advantages of combining CNNs and Transformers, they are designed as sequential models, where the architectures are connected in a linear order. The sequential approach often causes one model to dominate the decision-making process, as most computational resources or parameters are concentrated in that model, overshadowing the strengths of the other. Consequently, the performance does not surpass that of single-architecture models and fails to fully leverage both the microscopic and macroscopic levels. To address this limitation, we adopt a parallel architecture that simultaneously utilizes CNNs and Transformers. This design effectively orchestrates the strengths of both approaches, specifically targeting the capture of artifacts at the mesoscopic level.

Building on this, shallow feature maps provide microscopic features, while deep feature maps offer macroscopic semantics. Thus, using a multi-scale approach in the decoder stage to simultaneously decode feature maps at different scales can help the model more precisely capture artifacts at the mesoscopic level. Similarly, some existing models, such as MVSS-Net~\cite{MVSS_2021} and Trufor~\cite{trufor2023}, also utilize multi-scale decoding methods at this stage. However, these models assume equal weighting across all scales without explicitly adjusting the weights, potentially overlooking the differences between scales. This can lead to insufficient utilization of key features or overemphasis on less important features, negatively impacting overall performance. To effectively address this issue, we introduce an adaptive weighting module that dynamically adjusts the importance of each scale. Additionally, by pruning less significant scales, the model significantly reduces computational costs and parameters with minimal impact on performance.

In summary, we propose the Mesorch architecture, a hybrid model combining CNNs and Transformers that dynamically adjusts scale weights to efficiently represent the mesoscopic level. By pruning low-significance scales, we reduce parameters and computational costs, resulting in two baseline models. Testing on four datasets demonstrates SOTA performance in F1 score, robustness, and FLOPs.

Our contributions are threefold:
\begin{itemize}
\item We introduce the Mesorch architecture, a hybrid model that leverages the strengths of CNNs and Transformers in parallel. This architecture combines a multi-scale approach to effectively orchestrate microscopic and macroscopic levels, thereby precisely capturing mesoscopic level artifacts in IML tasks.
\item We propose an adaptive weighting module that dynamically adjusts the importance of each scale. Additionally, by selectively pruning less impactful scales, our approach considerably reduces computational costs and parameters while maintaining both robustness and performance.
\item We develop two baseline models based on the Mesorch architecture. Comprehensive experiments on benchmark datasets demonstrate that our method achieves SOTA performance in F1 score, robustness, and FLOPs.

\end{itemize}

\section{Related Work}
\subsection{Architectures in manipulation localization}

\paragraph{CNN-based Models}
In the realm of image manipulation localization, CNN-based models have long been the mainstream due to their robust feature extraction capabilities, excelling in capturing local textural anomalies indicative of manipulation. Some models like ManTra-Net~\cite{Mantra_2019} and SPAN~\cite{SPAN_2020} have been the mainstay in image manipulation localization, utilizing architectures such as VGG~\cite{VGG_2015} and ResNet-50~\cite{Resnet_2016} to effectively capture local textural anomalies. Recent advancements include integrating contrastive learning, as seen in models like NCL~\cite{NCL_IML_2023}, enhancing localization capabilities.

\paragraph{Transformer-based Models}
Building on the need for broader contextual understanding, Transformer-based models  Iml-vit~\cite{ma2023iml} is the first to integrate Transformer architectures into the IML domain, TruFor~\cite{trufor2023} utilize architectures like  SegFormer~\cite{SegFormer_2021}. These models are adept at synthesizing wide-ranging contextual information, dynamically focusing on areas of potential manipulation to improve localization accuracy, thus marking a significant advancement in the field. Similarly, MGQFormer~\cite{zeng2024mgqformer} uses a query-based Transformer architecture to pinpoint potential manipulation areas. Focusing on contextually relevant features refines manipulation localization precision, illustrating the potent impact of Transformer technologies in delivering robust, context-aware solutions that surpass prior methods.
\paragraph{Hybrid CNN-Transformer Models}
In addition to the aforementioned models, hybrid approaches that sequentially combine CNNs and Transformers, such as ObjectFormer~\cite{objectformer_2022}, have also gained significant attention. ObjectFormer employs the EfficientNet architecture as the initial part of its encoder to downsample input data into specific feature blocks, which are then passed sequentially into a dual-stream vision transformer. 

\subsection{Multi-Scale Applications}
In the decoder stage, multi-scale techniques have been adopted in image manipulation localization to enhance feature analysis at different resolutions. For example, MVSS-Net~\cite{MVSS_2021} utilizes a multi-view, multi-scale supervision strategy to detect image manipulations. By incorporating both local edge information and holistic context, MVSS-Net leverages multi-scale feature learning to accurately identify manipulated regions, offering a highly generalizable solution across various datasets and scenarios. Similarly, TruFor~\cite{trufor2023} employs the SegFormer~\cite{SegFormer_2021} backbone to effectively integrate spatial relationships across the entire image. Through its multi-scale approach, TruFor enhances the localization of manipulation inconsistencies by combining fine-grained local features with broad contextual information, ensuring robust and accurate manipulation localization.

\begin{figure*}[h]
    \centering
    \includegraphics[width=0.95\textwidth]{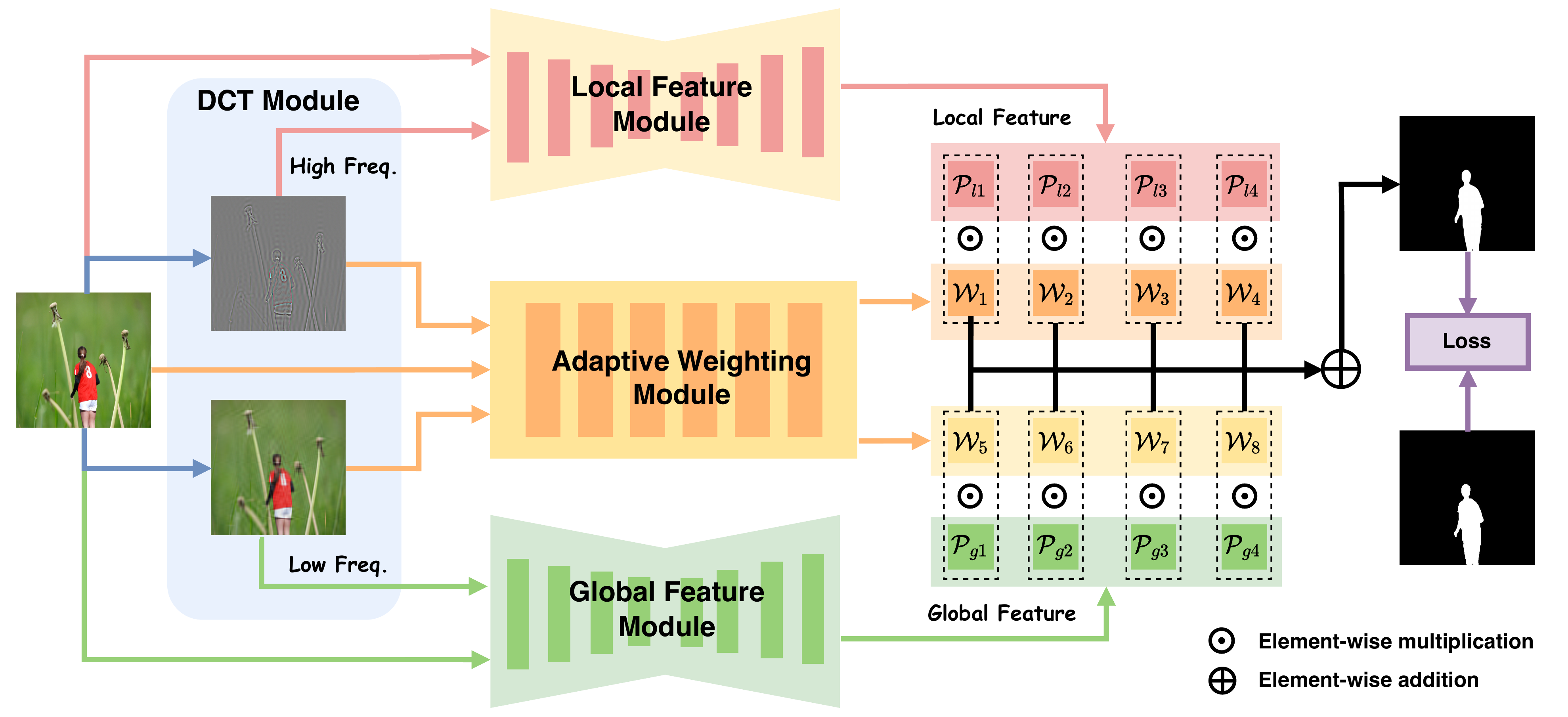}
    \caption{
Mesorch Framework: The input RGB image undergoes high- and low-frequency processing in the DCT Module to generate respective high-frequency and low-frequency representations. The Local Feature Module focuses on detecting fine-grained manipulation using both the original and high-frequency images, while the Global Feature Module captures object-level tampering cues by leveraging the original and low-frequency images. The Adaptive Weighting Module dynamically integrates these images by assigning pixel-level weights to local and global features. The final combined features are used for prediction and compared with ground-truth labels to compute the loss.}
    \label{fig:overview}
\end{figure*}

\section{Method}
In this section, we introduce the Mesorch framework, shown in Fig.\ref{fig:overview}. The process begins with an RGB image, which is processed to extract high- and low-frequency features using DCT~\cite{Digital_Image_Processingl_2018}. These features are then merged with the original image to create the high-frequency and low-frequency enhanced images. The high-frequency enhanced image is passed to the LocalFeatureModule, while the low-frequency enhanced image is passed to the GlobalFeatureModule. Each module outputs feature maps at four distinct scales. The corresponding decoder processes these feature maps to generate initial predictions of manipulated regions. These multi-scale predictions are weighted and combined to produce the final prediction. To further enhance efficiency, a pruning method is applied after initial model convergence, removing less significant scales to optimize the parameter count and FLOPs.

\subsection{Integration of Local and Global Information for Enhanced Prediction}
Artifacts hidden at the mesoscopic level may not be prominent in the RGB domain, but they can be amplified through frequency domain information~\cite{objectformer_2022}. Therefore, we extract high-frequency and low-frequency features from the frequency domain to enhance the capabilities of the local and global feature encoder modules.
\paragraph{Feature Enhancement with Discrete Cosine Transform}
Initially, as shown in Fig. \ref{fig:overview}(a), an RGB image \(\mathbf{x}\) with size \(H \times W \times 3\) is processed using the Discrete Cosine Transform (DCT) to separate it into high-frequency components \(\mathbf{x}_h\) and low-frequency components \(\mathbf{x}_l\). These components, retaining the dimension \( H \times W \times 3\), are then combined with the original image \(\mathbf{x}\) to form enhanced representations for further processing:

\begin{equation}
\mathbf{I}_h = \{\mathbf{x}, \mathbf{x}_h\}, \quad \mathbf{I}_h \in \mathbb{R}^{H \times W \times 6}
\end{equation}
\begin{equation}
\mathbf{I}_l = \{\mathbf{x}, \mathbf{x}_l\}, \quad \mathbf{I}_l \in \mathbb{R}^{H \times W \times 6}
\end{equation}

\paragraph{Feature Encoding and Scale-wise Decoding}
After further enhancing the high-frequency and low-frequency features, the high-frequency enhanced image \(\mathbf{I}_h\) is processed by the LocalFeatureEncoder, while the low-frequency enhanced image \(\mathbf{I}_l\) is processed by the GlobalFeatureEncoder. Each encoder subsequently outputs features at four distinct scales:

\begin{equation}
\begin{aligned}
\{L_{s_1}, L_{s_2}, L_{s_3}, L_{s_4}\} &= \text{LocalFeatureEncoder}(\mathbf{I}_h), \\
L_{s_i} &\in \mathbb{R}^{\frac{H}{2^{(i+1)}} \times \frac{W}{2^{(i+1)}} \times C_{i_{\text{local}}}}
\end{aligned}
\end{equation}

\begin{equation}
\begin{aligned}
\{G_{s_1}, G_{s_2}, G_{s_3}, G_{s_4}\} &= \text{GlobalFeatureEncoder}(\mathbf{I}_l), \\
G_{s_i} &\in \mathbb{R}^{\frac{H}{2^{(i+1)}} \times \frac{W}{2^{(i+1)}} \times C_{i_{\text{global}}}}
\end{aligned}
\end{equation}

Here, \(C_{i_{\text{local}}}\) and \(C_{i_{\text{global}}}\) denote the total number of output channels at each scale \(i\) for the local and global encoders, respectively.

The feature maps generated by the local and global feature encoder at scales \(i = 1, 2, 3, 4\) are then processed through the decoder, which outputs prediction masks with a shape of \(\frac{H}{4} \times \frac{W}{4} \times 1\) for each scale \(i\):

\begin{equation}
P_{l_i} = \text{LocalFeatureDecoder}(L_{s_i})
\end{equation}

\begin{equation}
P_{g_i} = \text{GlobalFeatureDecoder}(G_{s_i})
\end{equation}

The local and global predictions are combined to produce a summed final prediction mask \(P_{\text{summed}}\) with a shape of \(\frac{H}{4} \times \frac{W}{4} \times 1\). This mask is then resized to the original dimensions of the image (\(H \times W\)) to generate the final prediction mask \(P_{\text{final}}\):

\begin{equation}
P_{\text{final}} = \text{Resize}\left(\sum_{i=1}^{4} (P_{l_i} + P_{g_i}), H, W\right)
\end{equation}

The model's performance is measured by calculating the cross-entropy loss between \(P_{\text{final}}\) and the ground truth mask, which identifies the actually manipulated regions in the image. The ground truth mask serves as a binary map highlighting the manipulated areas, guiding the model to focus on discrepancies between the predicted and actual manipulations:

\begin{equation}
\text{Loss} = \text{CrossEntropyLoss}(P_{\text{final}}, \text{Mask})
\end{equation}

\subsection{Adaptive Scale Weighting and Model Pruning}
To address the issue of equal weighting across scales, which can lead to inefficient feature utilization, and to reduce the parameter count in the hybrid model, we introduce the AdaptiveWeightingModule and model pruning method to improve prediction accuracy and computational efficiency.
\begin{table*}[ht]
\small 
\centering
\caption{Comparison of model performances using standard F1 and permute-F1 metrics, where models denoted with ``-P" were trained with pruning methods.}
\label{tab:performance}
\begin{tabular}{@{}ccccccccccc@{}}
\toprule
\multirow{2}{*}{Model}  & \multicolumn{5}{c}{F1} & \multicolumn{5}{c}{Permute F1} \\ 
& Coverage & Columbia & NIST16 & CASIAv1 & Avg.& Coverage & Columbia & NIST16 & CASIAv1& Avg. \\ \midrule
MVSS-Net  &0.4860  & 0.7399  & 0.3363  & 0.5832  & 0.5364  & 0.5172 & 0.7879 & 0.3775 & 0.6016 & 0.5711 \\
PSCC-Net  &0.4475 & 0.8841 & 0.3457 & 0.6304 & 0.5769 & 0.4930 & 0.8937 & 0.3944 & 0.6382 & 0.6048 \\
CAT-Net  &0.4273 & \underline{0.9150} & 0.2521 & 0.8081 & 0.6006 & 0.5165 & 0.9547 & 0.3316 & 0.8154 & 0.6546 \\
TruFor  &0.4573 & 0.8845 & 0.3480 & 0.8176 & 0.6269 & 0.5369 & 0.9547 & 0.4046 & 0.8340 & 0.6826 \\
Mesorch (ours)  &\textbf{0.5862} & 0.8903 & \textbf{0.3921} & \underline{0.8398} & \textbf{0.6771} & \textbf{0.6342} & \textbf{0.9708} & \textbf{0.4514} & \underline{0.8472} & \textbf{0.7259} \\
Mesorch-P (ours)  & \underline{0.5470} & \textbf{0.9224} & \underline{0.3888} & \textbf{0.8465} & \underline{0.6762} & \underline{0.6107} & \underline{0.9662} & \underline{0.4441} & \textbf{0.8564} & \underline{0.7194} \\ \bottomrule
\end{tabular}
\end{table*}



\begin{table*}[ht]
\small 

\centering
\caption{Robustness test. ``Avg.F1" represents the average F1 score on CASIAv1 across perturbation strengths.}
\label{tab:robust}
\begin{tabular}{@{}llllllllll@{}}
\toprule
\multirow{2}{*}{Perturbation} & \multirow{2}{*}{Model} & \multicolumn{7}{c}{Standard Deviations} & \multirow{2}{*}{Avg.F1} \\ \cmidrule(l){3-9} 
 & & None & 3 & 7 & 11 & 15 & 19 & 23 &  \\ \midrule
\multirow{6}{*}{\textbf{GaussNoise}} 
 & MVSS-Net & 0.5832 & 0.5824 & 0.5822 & 0.5764 & 0.5736 & 0.5620 & 0.5613 & 0.5744 \\
 & PSCC-Net & 0.6304 & 0.6127 &0.5752&0.5540 & 0.5402& 0.5232 & 0.5115 & 0.5639 \\
 & CAT-Net & 0.8081 & 0.7979 & 0.7883 & 0.7832 & 0.7720 & 0.7573 & 0.7551 &0.7802 \\
 & Trufor & 0.8208 & 0.7666 & 0.7378 & 0.7190 & 0.6947 & 0.6832 & 0.6780 & 0.7286 \\
 & Mesorch(ours) & 0.8398 & 0.8205 & 0.8050 & 0.7968 & 0.7887 & 0.7780 & 0.7696 & \textbf{0.7998}\\
 & Mesorch-P(ours) & 0.8465 & 0.8184 & 0.7872 & 0.7694 & 0.7636 & 0.7543 & 0.7501 & \underline{0.7842} \\ \bottomrule
\multicolumn{1}{l}{} &\multirow{2}{*}{Model} & \multicolumn{7}{c}{Kernel Size}  & \multirow{2}{*}{Avg.F1} \\ \cmidrule(l){3-9} 
 & & None & 3 & 7 & 11 & 15 & 19 & 23 \\ \midrule
\multirow{6}{*}{\textbf{GaussianBlur}} 
 & MVSS-Net & 0.5832 & 0.4587 & 0.3097 & 0.2369 & 0.1890 & 0.1571 & 0.1392 & 0.2962 \\
 & PSCC-Net & 0.6304 & 0.5410 & 0.4531 & 0.3156 & 0.1655 & 0.1140 & 0.0775 & 0.3282 \\
 & CAT-Net & 0.8081 & 0.7512 & 0.6532 & 0.5435 & 0.4337 & 0.3142 & 0.2142 & 0.5312 \\
 & Trufor & 0.8208 & 0.7508 & 0.6881 & 0.6032 & 0.4563 & 0.2741 & 0.1304 & 0.5320 \\
 & Mesorch(ours) & 0.8398 & 0.7898 & 0.7081 & 0.6277 & 0.5328 & 0.4193 & 0.2940 & \underline{0.6016} \\
 & Mesorch-P(ours) & 0.8465 & 0.7867 & 0.7179 & 0.6380 & 0.5262 & 0.4106 & 0.2967 & \textbf{0.6032} \\ \bottomrule
\multicolumn{1}{l}{} &\multirow{2}{*}{Model} & \multicolumn{7}{c}{Quality Factors}  & \multirow{2}{*}{Avg.F1} \\ \cmidrule(l){3-9}
& & None & 100 & 90 & 80 & 70 & 60 & 50 \\ \cmidrule(l){2-9} 
\multirow{6}{*}{\textbf{JpegCompression}} 
 & MVSS-Net & 0.5832 & 0.5695 & 0.5446 & 0.5170 & 0.4906 & 0.4489 & 0.3888 & 0.5061 \\
 & PSCC-Net & 0.6304 & 0.6220 & 0.5789 & 0.4930 & 0.4518 & 0.3846 & 0.2869 & 0.4925 \\
 & CAT-Net & 0.8081 & 0.7896 & 0.7858 & 0.7431 & 0.7230 & 0.6838 & 0.6132 & 0.7352 \\
 & Trufor & 0.8208 & 0.8060 & 0.7938 & 0.7017 & 0.6852 & 0.6329 & 0.4942 & 0.7049 \\
 & Mesorch(ours) & 0.8398 & 0.8312 & 0.8194 & 0.7716 & 0.7706 & 0.7285 & 0.6552 & \textbf{0.7738} \\
 & Mesorch-P(ours) & 0.8465 & 0.8314 & 0.8219 & 0.7653 & 0.7598 & 0.7161 & 0.6239 & \underline{0.7664} \\ \bottomrule
\end{tabular}
\end{table*}

\paragraph{Scale-wise Importance}
The weighting network takes a concatenated input of the original RGB image \(\mathbf{x}\), its high-frequency component, and its low-frequency component, resulting in an input of size \(\mathbb{R}^{H \times W \times 9}\). The network then produces a normalized weight vector \(W\), where each element reflects the importance of each scale's prediction at every pixel:

\begin{equation}
\begin{aligned}
W &= \text{WeightingModule}(\{\mathbf{x}, \mathbf{x}_h, \mathbf{x}_l\}), \\
W &\in \mathbb{R}^{\frac{H}{4} \times \frac{W}{4} \times 8}
\end{aligned}
\end{equation}

\paragraph{Pixel-wise Prediction Fusion}
The final prediction of manipulated regions is determined by performing a weighted sum of the predicted masks across all scales. First, the combined mask \(P_{\text{all}}\) is formed by merging the local and global predicted masks from each scale, where \(P_{\text{all}} \in \mathbb{R}^{\frac{H}{4} \times \frac{W}{4} \times 8}\). Then, the weighted summation is calculated:

\begin{equation}
P_{\text{final}} = \text{Resize}\left(\sum_{j=1}^{8} W_j \cdot P_{all_j}, H, W\right)
\end{equation}

Here, \(P_{\text{final}}\) is the final prediction mask, resized to the original image dimensions (\(H \times W\)). As with the earlier prediction, this mask is compared to the ground truth mask to calculate the cross-entropy loss.

\paragraph{Rationale for Secondary Pruning}
Although the initial training phase allows the model to converge and identify potentially useful features across various scales, further analysis often reveals that some scales might contain redundant or even noisy information, which could diminish the model's overall effectiveness. Consequently, it becomes essential to evaluate the contribution of each scale using the following criteria:

First, the average weight \(\bar{W}_i\) for each scale \(i\) is calculated by averaging the weights \(W_{i,n}\) across all pixels or feature units \(n\) within that scale:

\begin{equation}
\bar{W}_i = \frac{1}{N} \sum_{n=1}^{N} W_{i,n}
\end{equation}

Next, the pruning condition is defined as follows:

\begin{equation}
\text{Prune Condition: } \bar{W}_i < \epsilon
\end{equation}

Here, \(\bar{W}_i\) represents the mean weight of the \(i\)-th scale, calculated over all \(N\) pixels units in that scale. If this average weight falls below a predefined threshold \(\epsilon\), the corresponding scale is deemed to contribute minimally to the model and is subsequently pruned. This approach ensures that the model focuses on scales that provide meaningful information while discarding those that add little to the overall performance.
.

\section{Experiments}
\subsection{Experimental Setup}
 \begin{figure*}[h]
    \centering
    \includegraphics[width=0.85\textwidth]{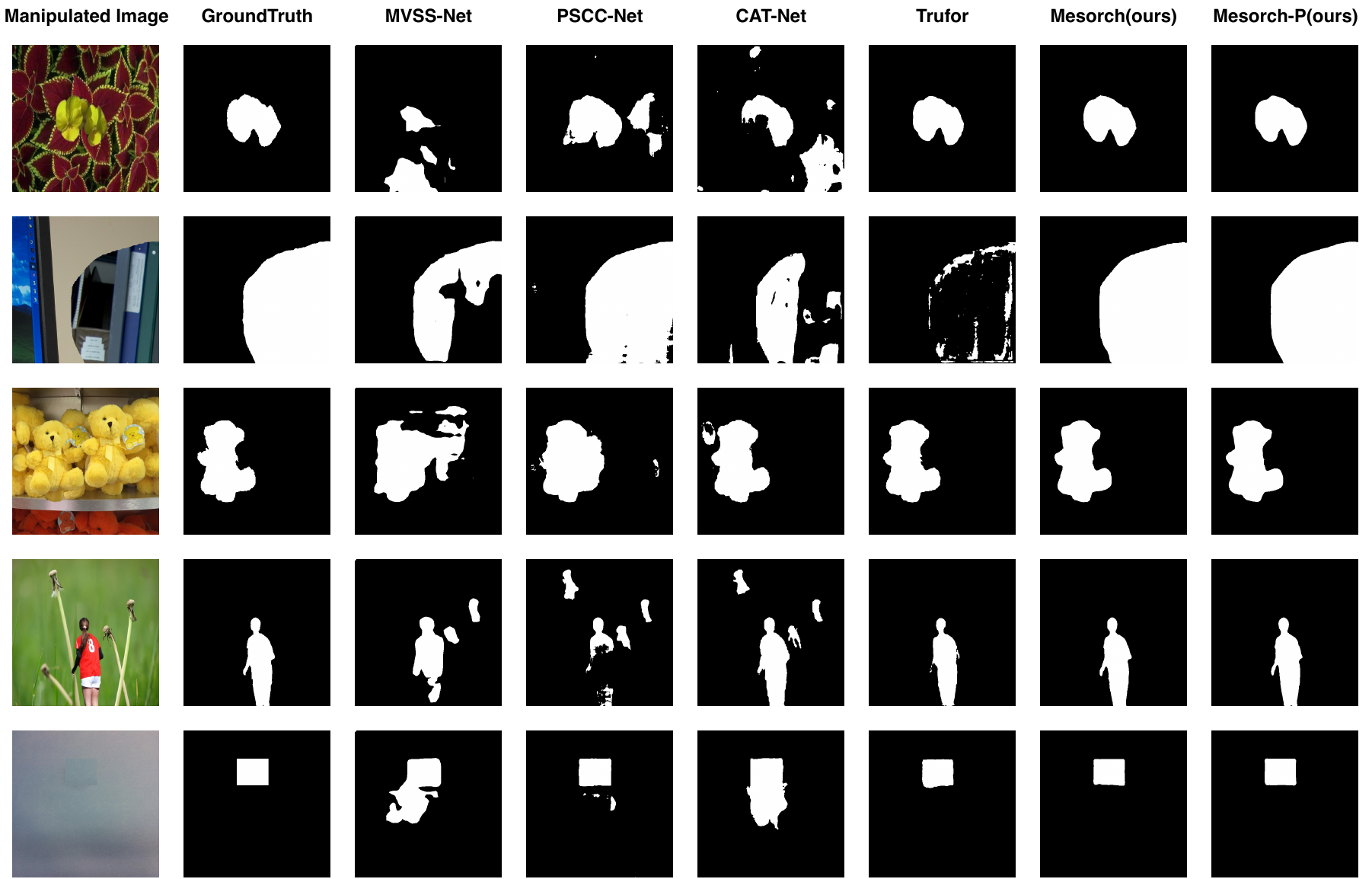}
    \caption{Qualitative analysis of SOTA models. We randomly selected and compared four semantically manipulated images and one non-semantically manipulated image based on their respective proportions in the datasets. The first four images are semantically manipulated, while the last one is non-semantically manipulated.}
    \label{fig:qualitative}
\end{figure*}

\paragraph{Training}Our model was trained using the standardized Protocol-CAT dataset, provided through a codebase referenced in~\cite{imdl2024}. This protocol includes established datasets and typical data augmentation methods. All images were resized to 512x512 pixels. We conducted the training over 150 epochs, utilizing a batch size of 12 on four NVIDIA 4090 graphics cards. The learning rate followed a cosine schedule~\cite{cosine_decay_2017}, starting at 1e-4 and tapering to a minimum of 5e-7, with a warm-up period of 2 epochs to gradually adjust the learning rate. The AdamW optimizer was used with a weight decay of 0.05 to mitigate overfitting. Furthermore, we set the accumulation iteration to 2, effectively adjusting the batch size to enhance the model’s generalization across diverse data inputs.

\paragraph{Testing} We conducted our model evaluations using publicly recognized benchmarks~\cite{imdl2024} across four widely used datasets: CASIA v1~\cite{CASIA_2013}, Coverage~\cite{Coverage_2016}, NIST16~\cite{NIST16_2019}, and Columbia~\cite{Columbia_2006}. These datasets are widely recognized for their diverse challenges and have been instrumental in assessing the generalization of image manipulation localization methods.

\paragraph{Metrics} As with our testing, we followed the same publicly recognized benchmarks for evaluation, employing standard pixel-level F1 scores to gauge performance. The results, calculated under the standard threshold of 0.5, provide a comprehensive assessment of localization accuracy.

\subsection{State-of-the-art comparison}
To ensure an accurate evaluation, we trained the models using open-source code on the resolutions recommended in their respective papers, employing the CAT-Net~\cite{CAT-Net2022} protocol dataset. We then benchmarked their performance across several well-established datasets using the F1 score. The comparative analysis included various methods such as PSCC-Net~\cite{liu2022pscc}, MVSS-Net~\cite{MVSS_2021}, CAT-Net~\cite{CAT-Net2022}, and Trufor~\cite{trufor2023}.

\paragraph{Localization results}In Table~\ref{tab:performance}, we have highlighted the best-performing model in bold and the second-best model with an underline across all evaluated datasets. Our proposed method, both before and after pruning, consistently ranks among the top performers, achieving either the highest or second-highest results. This consistent performance demonstrates the accuracy of our approach in effectively handling various image manipulation localization tasks. Additionally, Figure \ref{fig:qualitative} qualitatively shows that our model successfully captures both the object layout and fine details at the mesoscopic level, resulting in highly accurate manipulated masks.
\begin{table}[h]
\centering
\caption{Comparison of parameters and computational efficiency (Flops) across different models.}
\label{tab:para}
\begin{tabular}{@{}lcc@{}}
\toprule
Model & Parameters (M) & FLOPs (G) \\ \midrule
MVSS & 150.528 & 171.008 \\
PSCC & 3.668 & 376.832 \\
CAT-Net & 116.736 & 137.216 \\
TruFor & 68.697 & 236.544 \\
Mesorch(ours) & 85.754 & 124.928 \\
Mesorch-P(ours) & 62.235 & 64.821 \\
\bottomrule
\end{tabular}
\end{table}


\begin{table}[]
\centering
\caption{Independent Evaluation. ``Avg.F1" represents the average F1 score across four datasets.}
\label{tab:independent}
\begin{small}
\begin{tabular}{@{}llllll@{}}
\toprule
Model & Multi. & Weighting & Extractor & Prune & Avg.F1 \\ \midrule
\multirow{2}{*}{\textbf{CNN}} & $\times$ & $\times$ & $\times$ & $\times$ & 0.5610 \\ 
 & \checkmark & $\times$ & $\times$ & $\times$ & 0.6031 \\ \cmidrule(l){1-6}
\multirow{2}{*}{\textbf{Transformer}} & $\times$ & $\times$ & $\times$ & $\times$ & 0.5723 \\ 
 & \checkmark & $\times$ & $\times$ & $\times$ & 0.6096 \\ \cmidrule(l){1-6}
\multirow{6}{*}{\textbf{CNN+Trans.}} & $\times$ & $\times$ &  $\times$ & $\times$ & 0.6422 \\
& \checkmark & $\times$ & $\times$ & $\times$ & 0.6501 \\
& \checkmark & $\times$ & \checkmark & $\times$ & 0.6612 \\
& \checkmark & \checkmark & $\times$ & $\times$ & 0.6653 \\
& \checkmark & \checkmark & \checkmark & $\times$ & 0.6771 \\
& \checkmark & -- & \checkmark & \checkmark & 0.6762 \\ \bottomrule
\end{tabular}
\end{small}
\end{table}
\paragraph{Robust performance}To assess model robustness under various conditions, we tested on the CASIAv1 dataset and reported the results in Table~\ref{tab:robust}. We introduced Gaussian noise with varying standard deviations, Gaussian blur with different kernel sizes, and JPEG compression using various quality factors as perturbations. The results show that our model consistently achieved SOTA robustness across all three perturbations. Notably, even after pruning, our method maintained superior robustness compared to all previous models, demonstrating its effectiveness in handling diverse image distortions.

\begin{table}[]
\centering
\caption{The performance of the same module uses different backbones. ``Avg.F1" stands for ``Average F1," representing the mean value of the standard F1 score across four datasets.}
\label{tab:architecture}
\begin{tabular}{@{}lll@{}}
\toprule
CNN & Transformer & Avg.F1 \\ \midrule
\multirow{4}{*}{\textbf{ConvNeXt}} & MAE & 0.6215 \\
 & PVT & 0.6704 \\
 & Segformer & 0.6771 \\
 & Swin & 0.6089 \\ \cmidrule(l){1-3}
\multirow{4}{*}{\textbf{Resnet}} & MAE & 0.6327 \\
 & PVT & 0.6496 \\
 & Segformer & 0.6615 \\
 & Swin & 0.6022 \\ \bottomrule
\end{tabular}
\end{table}
\paragraph{FLOPs and Parameters} The number of parameters and FLOPs for all measurements were calculated based on a resolution of 512x512 and a batch size of 1. As shown in Table~\ref{tab:para}, our results demonstrate that our model has fewer FLOPs than all SOTA models, with a parameter count second only to PSCC-Net. Furthermore, the application of our pruning method further reduces FLOPs and total parameter count, giving our model a greater advantage over existing SOTA models.

\subsection{Ablation study}

\paragraph{Independent Evaluation of Proposed Method} To validate the effectiveness of our proposed method, we first independently evaluated the ConvNeXt architecture for CNNs and the Segformer architecture for Transformers as baseline models and assessed their performance under multi-scale conditions. After establishing the baseline performance, we further evaluated the hybrid model, gradually incorporating the multi-scale approach, weighting method, and DCT feature extraction.  We also tested the pruned model, as shown in Table~\ref{tab:independent}. The results demonstrate that each component proposed in the paper is crucial for accurately localizing image manipulations.

\begin{figure}[h]
    \centering
    \includegraphics[width=0.47\textwidth]{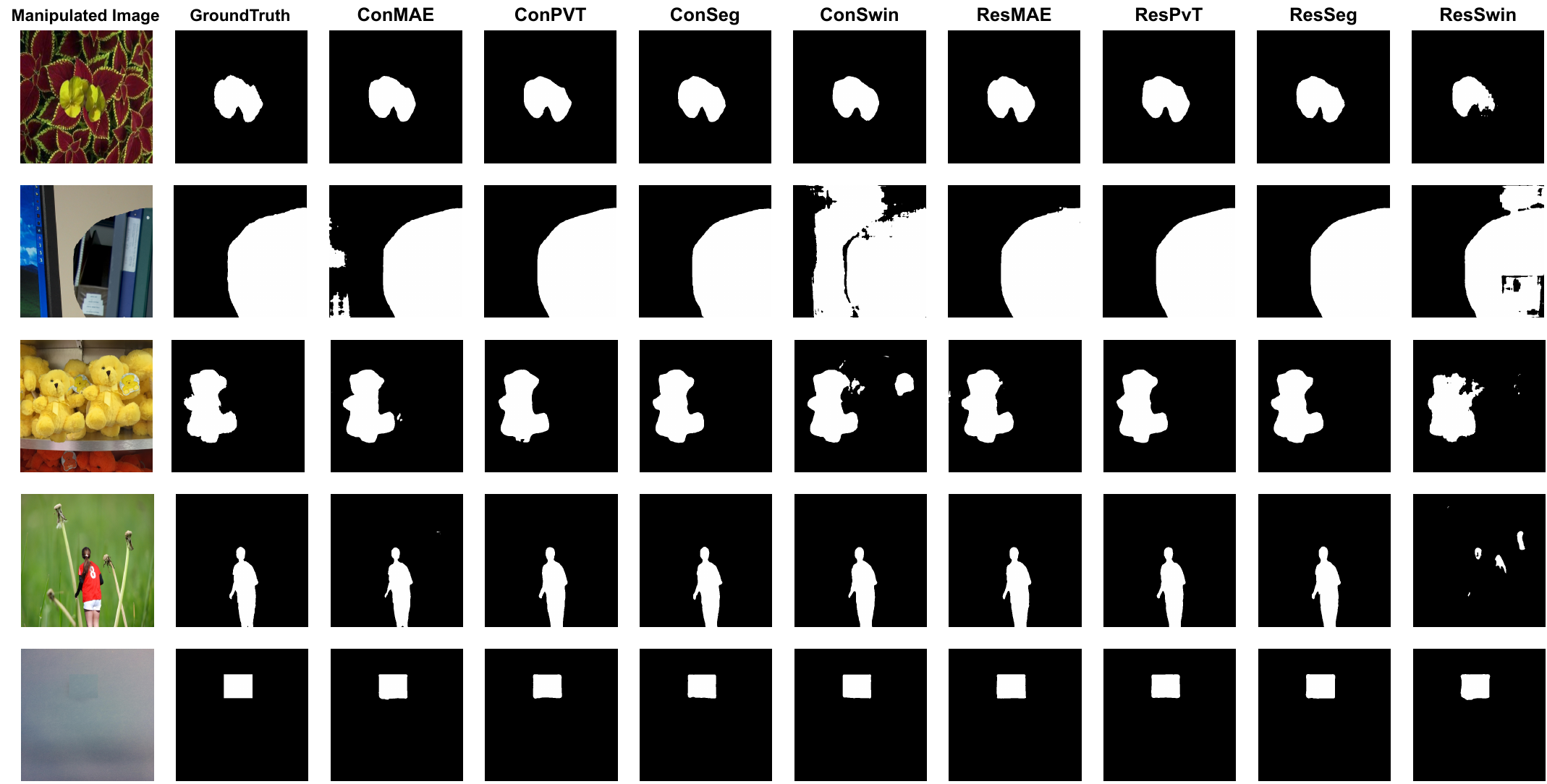}
    \caption{Ablation study on different backbones implemented in Mesorch Qualitatively ``Conv" is ConvNeXt, ``Res" is ResNet, and ``Seg" is Segformer.}
    \label{fig:long}
\end{figure}
\paragraph{Comparative Analysis of Model Architectures} We evaluated the performance of the Mesorch architecture using different combinations of two CNN models (Resnet-50~\cite{Resnet_2016} and ConvNeXt-Tiny~\cite{Convnet_2022}) with four Transformer models (MAE-Base~\cite{MAE_2022}, PvT-B3~\cite{pvtv22022}, Segformer-B3~\cite{SegFormer_2021}, and SwinTransformer-Base~\cite{Swin_2021}). The performance differences are summarized in Table \ref{tab:architecture}. Additionally, Figure \ref{fig:long} provides qualitative analysis, showcasing how the Mesorch architecture performs with different backbones. Our findings indicate that the combination of ConvNeXt and Segformer excels in both macroscopic localization and capturing microscopic features, outperforming other model combinations.

\section{Conclusion}
Inspired by the mesoscopic perspective, this paper redefines the IML task to orchestrate both microscopic and macroscopic levels. Building on this, we propose the Mesorch architecture, a hybrid model that leverages the strengths of CNNs and Transformers while dynamically adjusting scale weights to efficiently capture artifacts at the mesoscopic level. To reduce the parameter count and computational cost, we also introduce two baseline models based on this architecture. Extensive testing on large-scale datasets demonstrates that our approach consistently achieves SOTA performance in terms of F1 score, robustness, and FLOPs.

\section{Acknowledgments} 
The authors would like to express our gratitude to Professor Bin Li for introducing the mesoscopic concept in DiffForensics, which inspired our work. We also thank Xianhang Cheng and Kaiwen Feng for their contributions to the design of Fig. \ref{fig:overview}.

This work was jointly supported by the Sichuan Natural Science Foundation under grant 2024YFHZ0355, Sichuan Major Projects under grant 2024ZDZX0001, and the Science and Technology Development Fund, Macau SAR, under grants 0141/2023/RIA2 and 0193/2023/RIA3.  Numerical computations were jointly supported by Hefei Advanced Computing Center and Chengdu Haiguang Integrated Circuit Design Co., ltd. with HYGON K100AI DCU units.

\bibliography{aaai25}


\section{Appendix}

In Table~\ref{tab:auc} and Table~\ref{tab:iou}, the AUC and IOU metrics have been used to assess model performance across multiple datasets. The best-performing values are highlighted in bold, while the second-best values are underlined. Our proposed method, both with and without pruning, demonstrates SOTA performance, consistently achieving either the highest or second-highest scores in terms of AUC and IOU. This consistent high performance across various datasets underscores the effectiveness of our approach in accurately localizing image manipulations.

\begin{table}[H]
\small
\setlength{\tabcolsep}{1mm}
\centering
\caption{Comparison of model performances using AUC metrics, where models denoted with ``-P" were trained with pruning methods. Bold indicates the best value, and underlined indicates the second-best value.}
\label{tab:auc}
\begin{tabular}{@{}lllllllll@{}}
\toprule
Method & Coverage & Columbia & NIST16 & CASIAv1 & Avg.\\ \midrule
MVSS-Net & 0.8705 & 0.9332 & 0.7900 & 0.9115 & 0.8763 \\
CAT-Net & 0.9168 & 0.9457 & 0.8216 & 0.9804 & 0.9161 \\
PSCC-Net & 0.8838 & 0.9457 & 0.8279 & 0.9188 & 0.8941 \\
Trufor & 0.9423 & 0.8995 & 0.8781 & 0.9742 & 0.9235 \\
Mesorch & 0.9327 & 0.9094 & 0.8883 & 0.9852 & \underline{0.9289} \\
Mesorch-P & 0.9425 & 0.9447 & 0.9003 & 0.9831 & \textbf{0.9427} \\ \bottomrule
\end{tabular}
\end{table}

\begin{table}[H]
\small
\setlength{\tabcolsep}{1mm}
\centering
\caption{Comparison of model performances using IOU metrics, where models denoted with ``-P" were trained with pruning methods. Bold indicates the best value, and underlined indicates the second-best value.}
\label{tab:iou}

\begin{tabular}{@{}lllllllll@{}}
\toprule
Method & Coverage & Columbia &NIST16 & CASIAv1 & Avg. \\ \midrule
MVSS-Net & 0.3886 & 0.6721 & 0.2593 & 0.4907 & 0.4527 \\
CAT-Net & 0.3875 & 0.8953 & 0.2125 & 0.7481 & 0.5609 \\
PSCC-Net & 0.3009 & 0.8138 & 0.2971 & 0.4999 & 0.4779 \\
Trufor & 0.4149 & 0.8593 & 0.2964 & 0.7746 & 0.5863 \\
Mesorch & 0.5360 & 0.8766 & 0.3420 & 0.7877 & \textbf{0.6356} \\
Mesorch-P & 0.4989 & 0.9089 & 0.3367 & 0.7980 & \textbf{0.6356} \\ \bottomrule
\end{tabular}
\end{table}
\begin{figure*}[h]
    \centering
    \includegraphics[width=0.98\textwidth]{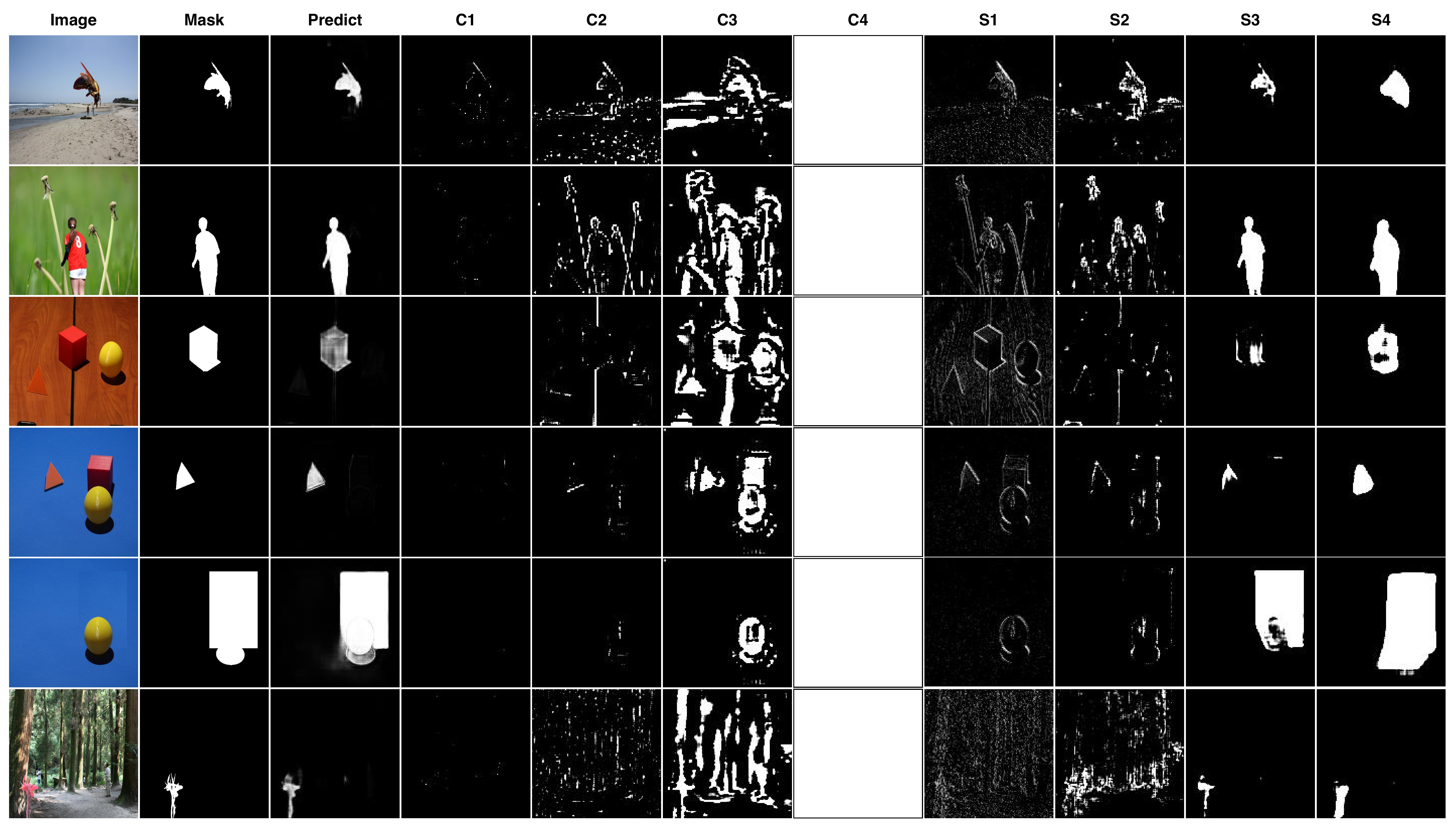}
    \caption{Comparisons of CNN and Transformer feature representations across different scales, demonstrating how CNNs excel at capturing local details in early layers but struggle to maintain structural coherence deeper into the network, while Transformers capture and preserve macro-level scene structure throughout their depths.}
    \label{fig:cnntrans}
\end{figure*}

Table \ref{tab:extractors} compares model performance across five different feature extractors, evaluated using the F1 score. The feature extractors, Bayar, Sobel, SRM, NoisePrint++, and DCT, were tested on four datasets. Columbia, NIST16, Coverage, and CASIAv1. Among these, the DCT feature extractor achieved the highest average performance with an F1 score of 0.6771, demonstrating its superior ability to extract discriminative features for this task. Notably, DCT outperformed other methods on the NIST16, Coverage, and CASIAv1 datasets, as highlighted by the bolded values in the table. This indicates that DCT is particularly effective in capturing the underlying patterns critical for this evaluation.

\begin{table}[H]
\centering
\small
\setlength{\tabcolsep}{1mm}
\caption{Comparison of Model Performance Using Different Feature Extractors. Bold Indicates the Best Value}
\label{tab:extractors}
\begin{tabular}{@{}llllll@{}}
\toprule
Extractor & Columbia & NIST16 & Coverage & CASIAv1 & Avg.F1 \\ \midrule
Bayar & 0.8973 & 0.3411 & 0.4282 & 0.7931 & 0.6149 \\
Sobel & 0.8718 & 0.2983 & 0.2943 & 0.7321 & 0.5491 \\
SRM & 0.8674 & 0.3417 & 0.5484 & 0.8358 & 0.6483 \\
NP++ & 0.9236 & 0.3046 & 0.5156 & 0.8394 & 0.6458 \\
DCT & 0.8903 & 0.3921 & 0.5862 & 0.8398 & \textbf{0.6771} \\ \bottomrule
\end{tabular}
\end{table}

Table \ref{tab:dct} shows the results of an ablation study comparing the performance of CNN and Transformer architectures using high-frequency and low-frequency DCT features across four datasets: Columbia, NIST16, Coverage, and CASIAv1. The study evaluates combinations of high-frequency and low-frequency DCT structures for both CNN and Transformer models. The results indicate that the “High-Frequency CNN + Low-Frequency Transformer” configuration achieves the best average performance, with an F1 score of 0.6771. This demonstrates the effectiveness of leveraging high-frequency features in CNNs and low-frequency features in Transformers, highlighting the complementary strengths of these architectures in capturing essential patterns in the data.

\begin{table}[H]
\centering
\small
\setlength{\tabcolsep}{1mm}
\caption{Ablation Study on CNN and Transformer Architectures with High-Frequency and Low-Frequency DCT Features}
\label{tab:dct}
\begin{tabular}{@{}lllllll@{}}
\toprule
CNN & Trans. & Columbia & NIST16 & Coverage & CASIAv1 & Avg.F1 \\ \midrule
High & Low & 0.8903 & 0.3921 & 0.5862 & 0.8398 & \textbf{0.6771} \\
High & High & 0.9191 & 0.3417 & 0.5245 & 0.8493 & 0.6587 \\
Low & High & 0.9049 & 0.3812 & 0.5544 & 0.8356 & 0.6690 \\
Low & Low & 0.8889 & 0.3448 & 0.4867 & 0.8305 & 0.6377 \\ \bottomrule
\end{tabular}
\end{table}

Fig.\ref{fig:cnntrans} presents a side-by-side examination of feature maps extracted at progressively deeper layers from a CNN-based model (C1–C4) and a Transformer-based model (S1–S4). The CNN’s early layers effectively capture fine-grained, local details, but as depth increases, the network increasingly loses track of global structures. In contrast, the Transformer acquires a strong sense of global composition even at shallow layers, and this coherence remains intact at deeper levels. These observations underscore the fundamental differences in how CNNs and Transformers learn and represent information at varying scales, providing empirical support for the design principles behind Mesorch.
\end{document}